# REDUCTION OF FEATURE VECTORS USING ROUGH SET THEORY FOR HUMAN FACE RECOGNITION

D. Bhattacharjee[*], D. K. Basu[**], M. Nasipuri[**], M. Kundu[**]

[*]Department of Computer Science and Engineering, University of Calcutta, Kolkata, India.

[**]Department of Computer Science and Engineering, Jadavpur University, Kolkata, India.

## ABSTRACT

In this paper we describe a procedure to reduce the size of the input feature vector. A complex pattern recognition problem like face recognition involves huge dimension of input feature vector. To reduce that dimension here we have used eigenspace projection (also called as Principal Component Analysis), which is basically transformation of space. To reduce further we have applied feature selection method to select indispensable features, which will remain in the final feature vectors. Features those are not selected are removed from the final feature vector considering them as redundant or superfluous. For selection of features we have used the concept of reduct and core from rough set theory. This method has shown very good performance. It is worth to mention that in some cases the recognition rate increases with the decrease in the feature vector dimension.

## 1. Introduction

In our day to day life, we find that the identification of some individual is done based on his/her face. This is the reason for which there had been so many suspense movies based on similar face concept. If human cloning is allowed in future then identification systems based on human face may become irrelevant. But till such happens, this can be a fertile research area for identification. There are other identification systems, based on retina, fingerprint, DNA etc., which are happened to be more accurate than face recognition, but involve the complexity that the person concerned has to be informed about the identification. This could be good for any security system, but may not be good for any investigative process. In such situations, face recognition plays an important role to identify any individual.

Like other complex pattern recognition applications face recognition system also encounters input feature vectors with larger dimensionality. Total number of training samples that is required to train a system is an exponential function of the dimension of the input feature vector. This is termed as "curse of dimensionality"[1], [20]. Due to the measurement cost and classification accuracy, the number of features should be kept as small as possible. A small and functional feature set makes the system work faster and use less memory. Dimensionality reduction is basically a space transformation where patterns from a higher dimension space is reduced and represented in a comparatively lower dimensional space. After dimensionality reduction some selected features constitute the final feature vector which is input to the recognition system. Feature selection in pattern recognition involves the derivation of certain features from the input data in order to reduce the amount of data used for classification. In both the cases i.e. dimensionality reduction and feature selection main objective is to reduce data without affecting the discriminatory power of the feature vectors. Dimensionality reduction is basically a transformation where selection keeps valuable data and removes redundant or superfluous data.

The organization of the article is as follows: Section 2 discusses some recent works on face recognition. Principal component analysis and eigenspace projection is discussed in section 3. Some feature selection techniques are given in Section 4. Section 5 provides a brief overview of rough set theory and relevant design details of feature selection using rough set theory. A brief discussion about Multi-layer Perceptron in given is Section 6. Finally, the effectiveness of the proposed model in terms of experimental results is provided in Section 7.

## 2. Literature Review

Zhang et al [5] have used Karhunen-Loève (KL) representation of faces, which is widely known as principal component analysis. However, the discriminatory power of KL representation mostly depends on the separation of different faces (different persons) than the spread (variance of faces along specific direction) of all faces of a particular class. If there are large variations in face images of a particular class, then the separation between different classes may decrease and as a consequence this method gives poor result for large variation in the face images. Atick et al's work [6] is a significant extension to the eigenface method. Its basic idea is to model the frontal (physical) human face as a surface in 3-D space and recover it from its 2-D image. They have suggested to expand the surface with respect to a set of eigensurfaces, called eigenheads. The significant advantage is that it is independent of the image-formation process (illumination, etc.). Hence, eigenhead-based face recognition could be significantly more robust. In Nearest Feature Line (NFL) method [7], any two feature-points of the same class (person) are generalized by the feature line (FL) passing through the two points. The derived FL can capture more variations of face images than the original points and thus expands the capacity of the available database. Moghaddam et al. [8] had proposed an interesting elastic matching technique where, the basic idea is to consider two images as 2-D surfaces in a 3-D space. A match is obtained by deforming one surface so that its shape becomes close to that of the other surface.

Teuvo Kohonen [9] demonstrated that a simple neural net could perform face recognition for aligned and normalized images of faces. Kohonen's system was not a practical success. Kung et al [10] proposes a neural-net technique and test it on a subset of the FERET database. The neural net considered by them, called decision-based neural net, classifies its input, which is basically a preprocessed and subsampled face images. Very high recognition rates were reported (96-99%). In another method, N. Intrator et al [11] have used feed-forward artificial neural network for feature extraction and classification, which is time-consuming (specially, in training). In another complex method, namely Convolutional Neural Network [12], [13], a Self-Organizing Map (SOM) or KL transform is used for dimensionality reduction and either Multi-layer Perceptron (MLP) or Nearest-Neighbor classifier is used for classification of images. The result obtained is obviously good. However, use of one neural network for dimensionality reduction and use of more than one neural networks for recognition i.e. convolution of neural networks is not necessarily be wise in terms of computation time, calculation complexity and easy understanding. The performance of autoassociation and classification nets, as reported in [5], is upper bounded by that of the eigenface but is more difficult to implement in practice. The artificial neural network Group-based Adaptive Tolerance (GAT) trees model [14] is used for translation-invariant face recognition, which was designed for airport security system. The Probabilistic Decision-Based Neural Network(PDBNN)[15] uses k different subnets for a k-people recognition problem. A subnet j in the PDBNN recognizer estimates the distribution of the pattern of person j only, and treats those patterns which do not belong to person j as the "non j" patterns. Face recognition based on extracted Gabor features classification using Multi-layer perceptron has been reported in [16]. In another method [17], face recognition is done using integration of two different fuzzy matching techniques. The first technique is based on the kernel-matching concept and second one uses local shift invariant Discrete Cosine Transform (DCT) coefficients of images. The final result is calculated by integrating the results obtained from both the techniques. The result of recognition on ORL face database is very high, where error rate is only 2.625%. RBF Neural Network [24] extracts face features first by PCA and then resulting features are projected into the Fisher's optimal subspace and finally, a hybrid learning algorithm is proposed to train the RBF neural networks. Bartlett et al [25] used a version of Independent Component Analysis (ICA) derived from the principle of optimal information transfer through sigmoidal neurons. Two different architectures were employed and combined result of them shown very good result on FERET face database.

Another method[26], kernel PCA uses kernel functions to achieve nonlinear mapping between the input space and the feature space. Therefore, the computation is a function of training examples rather than the dimension of the feature space. Discriminative common vectors method [27] is based on a variation of Fisher's Linear Discriminant Analysis for the small sample size case. This method is superior to other methods like Eigenface, Fisherface, and Direct-LDA. Another method, [28] nonlinear discriminant analysis involving a set of locally linear transformations yield locally linearly transformed classes that maximize the between-class covariance while minimizing within-class covariance. In modular PCA approach [29], the face images are divided into smaller sub-images and the PCA approach is applied to each of these sub-images. Since some of the local facial features of an individual do not vary even when the pose, lighting direction, and facial expression vary, it is experimentally confirmed that the modular PCA is able to cope with these variations. Another method [30], first derives a Gabor feature vector from a downsampled Gabor wavelet representations of face images, then reduces the dimensionality of the vector by means of PCA, and finally defines the independent Gabor features based on the independent component analysis. The independent property of these Gabor features facilitates the application of the probabilistic reasoning model for classification. Gabor-based kernel PCA method[31] integrates the Gabor wavelet representation of face images and the kernel PCA method for face recognition. Gabor wavelets first derive desirable facial features characterized by spatial frequency, spatial locality, and orientation selectivity to cope with the variations due to illumination and facial expression changes. The kernel PCA method is then extended to include fractional power polynomial models for enhanced face recognition performance.

## 3. Dimensionality Reduction

Any raw image data has many degrees of freedom and it is not possible to handle all those dimensions. To achieve an identification system, it is required that it gives result within a stipulated time. Therefore, such systems are designed to work in the domain of low dimension so that some result can be obtained when it is required. If the number of features, i.e. the pattern dimension, is more then the classifier will take more time to measure the intra-class and interclass distances. Moreover, if the pattern dimension is less for limited training samples, then the curse of dimensionality can be alleviated. On the other hand, reduction in the number of features may lead to a loss in the discrimination power and thereby lower the accuracy of the resulting recognition system. Feature extraction methods try to reduce the feature dimensions used in the classification step. There are especially two methods used in pattern recognition to reduce the feature dimensions; Principal Component Analysis (PCA) and Linear Discriminant Analysis (LDA). The relative places of the data between each other are never changed according to the utilized feature dimension reduction technique, PCA or LDA. Only the axes are changed to handle the data from a "better" point of view. Better point of view is simply generalization for PCA and discrimination for LDA. In this work we have considered Principal Component for dimensionality reduction.

### 3.1 Principal Component Analysis (PCA)

The advantage of PCA comes from its generalization ability. It reduces the feature space dimension by considering the variance of the input data. The method determines which projections are preferable for representing the structure of the input data. Those projections are selected in such a way that the maximum amount of information (i.e. maximum variance) is obtained in the smallest number of dimensions of feature space.

In order to obtain the best variance in the data, the data is projected to a subspace of the input image space, which is built by eigenvectors projected from data. In that sense, the eigenvalue corresponding to an eigenvector represents the amount of variance that eigenvector handles.

Eigenspace method is the implementation of Principal Component Analysis (PCA) over images. In this method, the features of the studied images are obtained by looking for the maximum deviation of each image from the mean image. This variance is obtained by getting the eigenvectors of the covariance matrix of all the images.

An image can be thought as a point in the image space by converting the image to a long vector by concatenating each column of the image one after the other.

When all the face images are converted into vectors, they will group at a certain location in the image space as they have similar structure, having eye, nose and mouth in common and their relative position correlated. This correlation is the main point to start the eigenspace analysis.

The Eigenspace method tries to find a lower dimensional space for the representation of the face images by eliminating the variance due to non-face images; that is, it tries to focus on the variation just coming out of the variation between the face images.

### 3.2 Eigenspace Projection

Eigenspace is calculated by identifying the eigenvectors of the covariance matrix derived from a set of training images. The eigenvectors corresponding to non-zero eigenvalues of the covariance matrix form an orthonormal basis that rotates and/or reflects the images in the N-dimensional space. Specifically, each image is stored in a vector of size N. Eigenspace projection involves following steps.

(a) The images are mean centered by subtracting the mean image from each image vector.
(b) These vectors are combined, side-by-side, to create a data matrix of size NxP (where P is the number of images).
(c) The data matrix X is multiplied by its transpose to calculate the covariance matrix.

(d) This covariance matrix has up to Q eigenvectors associated with non-zero eigenvalues, assuming Q<N. The eigenvectors are sorted, high to low, according to their associated eigenvalues. The eigenvector associated with the largest eigenvalue is the eigenvector that finds the greatest variance in the images. The eigenvector associated with the second largest eigenvalue is the eigenvector that finds the second most variance in the images. This trend continues until the smallest eigenvalue is associated with the eigenvector that finds the least variance in the images.

### 3.3 Algorithm for dimensionality reduction using Eigenspace projection

The following steps create an eigenspace.

**Step1)** **Center data:** Each of the training images must be centered. Subtracting the mean image from each of the training images centers the training images as shown in equation (1). The mean image is a column vector such that each entry is the mean of all corresponding pixels of the training images.

$$\overline{x}^i = x^i - m, \quad where \ m = \tfrac{1}{P}\sum_{i=1}^{P} x^i \ and \ x^i = \left[x_1^i, x_2^i, \ldots, x_N^i \right] \tag{1}$$

**Step2)** **Create data matrix:** Once the training images are centered, they are combined into a data matrix of size NxP, where P is the number of training images and each column is a single image as shown in equation (2).

$$X = \begin{bmatrix} \overline{x}_1^1 & \overline{x}_1^2 & \cdots & \overline{x}_1^P \\ \overline{x}_2^1 & \overline{x}_2^2 & \cdots & \overline{x}_2^P \\ \vdots & \vdots & \ddots & \vdots \\ \overline{x}_N^1 & \overline{x}_N^2 & \cdots & \overline{x}_N^P \end{bmatrix} \tag{2}$$

**Step3)** **Create covariance matrix:** The data matrix is multiplied by its transpose to create a covariance matrix as shown in equation (3).
$$C = X X^T \tag{3}$$

**Step4)** **Compute the eigenvalues and eigenvectors:** The eigenvalues and corresponding eigenvectors are computed for the covariance matrix.
$$CW = \Lambda W \tag{4}$$

here W is the set of eigenvectors associated with the eigenvalues $\Lambda$.

**Step5)** **Order eigenvectors:** Order the eigenvectors $w_i \in W$ according to their corresponding eigenvalues $\lambda_i \in \Lambda$ from high to low. Keep only the eigenvectors associated with non-zero eigenvalues. This matrix of eigenvectors is the eigenspace $W$, where each column of W is an eigenvector.

$$W = [w_1, w_2, \ldots, w_Q] \tag{5}$$

### 4. FEATURE SELECTION

Until this point, when creating a subspace using eigenspace projection we use all eigenvectors associated with non-zero eigenvalues. The computation time of eigenspace projection is directly proportional to the number of eigenvectors used to create the eigenspace. Therefore, if it is possible to remove some of the eigenvectors then computation time would decrease. Furthermore, by removing additional eigenvectors that do not contribute to the classification of the image, performance can be improved. Many variations of eigenvector selection exist. Five of them are discussed below.

i) **Standard eigenspace projection:** All eigenvectors corresponding to non-zero eigenvalues are used to create the subspace.
ii) **Remove the last 40% of the eigenvectors:** Since the eigenvectors are sorted by the

corresponding descending eigenvalues, this method removes the eigenvectors that find the least amount of variance among the images. Specifically, 40% of the eigenvectors that find the least amount of variance are removed [3].

**iii) Energy dimension:** Rather than use a standard cutoff for all subspaces, this method uses the minimum number of eigenvectors to guarantee that energy (*e*) is greater than a threshold. A typical threshold is 0.9. The energy of the $i^{th}$ eigenvector is the ratio of the sum of the first *i* eigenvalues over the sum of all the eigenvalues [2]

$$e_i = \frac{\sum_{j=1}^{i} \lambda_j}{\sum_{j=1}^{Q} \lambda_j} \tag{6}$$

**iv) Stretching dimension:** Another method of selecting eigenvectors based on the information provided by the eigenvalues is to calculate the stretch (s) of an eigenvector. The stretch of the $i^{th}$ eigenvector is the ratio of the $i^{th}$ eigenvalue ($\lambda_i$) over the maximum eigenvalue ($\lambda_1$) [2]. A common threshold for the stretching dimension is 0.01.

$$s_i = \frac{\lambda_i}{\lambda_1} \tag{7}$$

**v) Removing the first eigenvector:** The previous three methods assume that the information in the last eigenvectors work against classification. This method assumes that information in the first eigenvector works against classification. For example, lighting causes considerable variation in otherwise identical images. Hence, this method removes the first eigenvector [3].

## 5. Feature Selection using Rough Set

The optimality of eigenspace projection is based on the minimization of error to represent face images through individual principal components. Therefore, eigenspace projection does not guarantee that the reduced space with transformed feature vectors is minimal. That means further reduction of the transformation space can not be ruled out. Now to keep discriminative features from the principal components rough set theory can be applied. At first rough set theory introduced in brief and then the algorithm for feature selection has been discussed.

### 5.1 Rough Set

Rough set theory proposed by Professor Z. Pawlak [4]. Rough set theory has many applications in the areas like soft computing, machine learning, knowledge representation, decision making, data mining, expert systems, pattern classification and scene analysis. Rough set theory has sufficient advantages over statistical and probabilistic reasoning, as it does not require any additional information about data a priori.

To define the rough set let us consider a knowledge base K=(U, R) where U is the set of objects of our interest called the *universe* and an indiscernibility relation $R \subseteq U \times U$, is an equivalence relation. Any set X, which is a subset of U, can be characterized with respect to *R* as follows:

• The R-*lower approximation* of *X* is the set of all objects, which can be certainly classified as *X* with respect to *R* and that can be given as

$$R_*(X) = \bigcup \{Y \mid Y \in U/R \text{ and } Y \subseteq X\} \tag{8}$$

where, U/R means the family of equivalence classes of R (or classification of U).

• The R-*upper approximation* of *X* is the set of all objects, which can be *possibly* classified as *X* with respect to *R* and that can be given as

$$R^*(X) = \bigcup \{Y \mid Y \in U/R \text{ and } Y \cap X \neq \emptyset\} \tag{9}$$

●The R-*boundary region* of a set *X* with respect to *R* is the set of all objects, which can be classified neither as *X* nor as not-*X* with respect to *R* and that is given as

$$RN_R(X) = R^*(X) - R_*(X) \qquad (10)$$

From the above definitions it is evident that it if the boundary region is empty then the set *X* is crisp i.e. exact with respect to *R* but if the boundary region is nonempty then the set *X* is rough i.e. inexact with respect to *R*.

**Example 1** : To illustrate the concept let us consider a knowledge base K=(U, R), where U={$x_1$, $x_2$, ….., $x_8$}, and an equivalence relation R with the following equivalence classes:

$E_1$ = { $x_1$, $x_4$, $x_8$ }
$E_2$ = { $x_2$, $x_5$, $x_7$ }
$E_3$ = { $x_3$ }
$E_4$ = { $x_6$ }

For the set X = { $x_3$, $x_6$, $x_8$ }, we get

$R_*(X)$ = $E_3 \cup E_4$ = { $x_3$, $x_6$ }
$R^*(X)$ = $E_1 \cup E_3 \cup E_4$ = { $x_1$, $x_3$, $x_4$, $x_6$, $x_8$ }
$RN_R(X)$ = $E_1$ = { $x_1$, $x_4$, $x_8$ }

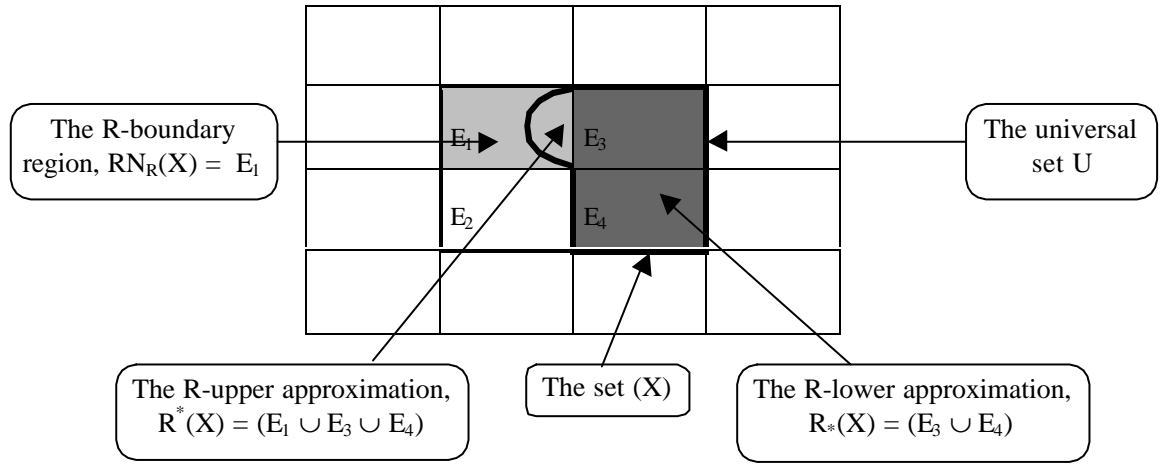

The R-boundary region, $RN_R(X) = E_1$

The universal set U

The R-upper approximation, $R^*(X) = (E_1 \cup E_3 \cup E_4)$

The set (X)

The R-lower approximation, $R_*(X) = (E_3 \cup E_4)$

**Figure 1**

### 5.1.1 Decision Tables

A decision table consists of two different attribute sets. One attribute set is designated to represents conditions(C) and another set is to represent decision(D). Therefore, each row of a decision table describes a decision rule, which indicates a particular decision to be taken if its corresponding condition is satisfied.

If a set of decision rules has common condition but different decisions then all the decision rules belonging to this set are inconsistent decisions, otherwise; they are consistent.

### 5.1.2 Dependency of Attributes

Similar to relational databases, dependencies between attributes may be discovered. If all the values of attributes from D are uniquely determined by values of attributes from C then D depends totally on C or C functionally determines D which is denoted by C $\Rightarrow$ D. If D depends on some of the attributes of C (i.e. not on all) then it is a partial dependency C $\Rightarrow_k$ D and a degree of dependency(k; $0 \le k \le 1$) can be computed as k = $\gamma$(C, D), where $\gamma$(C, D) is the consistency factor of the decision table. $\gamma$(C, D) is defined as the ratio of the number of consistent decision rules to the total number of decision rules in the decision tables.

If D depends in degree k, $0 \le k \le 1$, on C, then

$$g(C,D) = \frac{|POS_C(D)|}{|U|}, \tag{11}$$

where

$$POS_C(D) = \bigcup C_*(X), X \in U/I(D) \tag{12}$$

and I(D) is the indiscernibility relation over D, given as $\subseteq$ D.

The expression $POS_C(D)$, called C-positive region which contains all the elements of U that can be classified into distinct classes defined by I(D), based on information available in I(C).

**5.1.3 Reduction of Attributes**

Decision tables where feature vectors are the condition (C) and desired values for corresponding classes are the decisions(D) can also represent classification of feature vectors. Now the dimensionality reduction can simply be considered as removal of some attributes from the decision table(actually some features from the feature vector) preserving its basic classification capability. In this case, we would face two questions:
  i) whether a decision table contains some redundant or superfluous data, if yes then
  ii) collect those redundant data and remove them.

To perform such reduction of attributes in rough set theory we select indispensable attributes. In rough set theory indispensable attributes are selected using two fundamental concepts : reduct and core. They are defined as follows.

Let Q be a subset of P and let *a* belong to Q.
- *a* is *dispensable* in Q if $I(Q) = I(Q - \{a\})$; otherwise *a* is *indispensable* in Q.
- Set Q is *independent* if all its attributes are indispensable.
- Subset Q' of Q is a reduct of Q if Q' is independent and $I(Q') = I(Q)$.

Thus a reduct is the minimal subset of attributes that enables the same classification of elements of the universe as the whole set of attributes i.e., attributes that do not belong to a reduct are redundant or superfluous. These attributes can easily be removed or eliminated without affecting the classification of elements of the universe.

The core of a set T is the set of all indispensable attributes of T, which can be given as

$$CORE(T) = \bigcap RED(T), \tag{13}$$

where RED(T) is the set of all reducts of T.

Since the core is the intersection of all reducts, no element of the core can be eliminated affecting the classification power of attributes.

If for a dependency $C \Rightarrow D$, *D* depends on E where $E \subseteq C$ then E is called as relative D-reduct of C. Relative *D-core* of *C* is given by

$$CORE_D(C) = \bigcap RED_D(C) \tag{14}$$

where $Red_D(C)$ is the family of all *D*-reducts of *C*.

**Example2**: Let use consider the following example, given by Pawlak[4] to illustrate the feature selection procedure using reduct and core from rough set theory.

Suppose we are given the following decision table, the Table 1, where a, b, and c are condition attributes and d is a decision attribute. Here, our objective is to remove superfluous attributes in every decision rule. In order to achieve this first we have to compute core values of condition attributes in every decision rule. For the sake of illustration, let us compute core values of condition attributes for the first decision rule, i.e. the core family of sets F = $\{[1]_a, [1]_b, [1]_c\}$ = $\{\{1, 2, 4, 5\}, \{1, 2, 3\}, \{1, 4\}\}$, where $[x]_R$ denotes a category in R containing an element x $\in$ U and a(1) = 1, b(1) = 0, and d(1) = 1. In order to find dispensable categories, we have to drop one category at a time and check whether the intersection of remaining categories is still included in the decision category $[1]_d$ = $\{1, 2\}$, i.e.

$[1]_b \cap [1]_c = \{1, 2, 3\} \cap \{1, 4\} = \{1\}$
$[1]_a \cap [1]_c = \{1, 2, 4, 5\} \cap \{1, 4\} = \{1, 4\}$
$[1]_a \cap [1]_b = \{1, 2, 4, 5\} \cap \{1, 2, 3\} = \{1, 2\}$

**Table 1**

| U | a | b | c | d |
|---|---|---|---|---|
| 1 | 1 | 0 | 1 | 1 |
| 2 | 1 | 0 | 0 | 1 |
| 3 | 0 | 0 | 0 | 0 |
| 4 | 1 | 1 | 1 | 0 |
| 5 | 1 | 1 | 2 | 2 |
| 6 | 2 | 1 | 2 | 2 |
| 7 | 2 | 2 | 2 | 2 |

This means that the core value is $b(1) = 0$. Similarly we can compute remaining core values of condition attributes in every decision rule and the final results are given in Table 2 below.

**Table 2**

| U | a | b | c | d |
|---|---|---|---|---|
| 1 | - | 0 | - | 1 |
| 2 | 1 | - | - | 1 |
| 3 | 0 | - | - | 0 |
| 4 | - | 1 | 1 | 0 |
| 5 | - | - | 2 | 2 |
| 6 | - | - | - | 2 |
| 7 | - | - | - | 2 |

After computation of core values now we are in a position to compute value reducts. To compute reducts of the family $F = \{[1]_a, [1]_b, [1]_c\} = \{\{1, 2, 4, 5\}, \{1, 2, 3\}, \{1, 4\}\}$ we have to find all subfamilies $G \subseteq F$ such that $\cap G \subseteq [1]_d = \{1, 2\}$. There are three following subfamilies of F

$[1]_b \cap [1]_c = \{1, 2, 3\} \cap \{1, 4\} = \{1\}$
$[1]_a \cap [1]_c = \{1, 2, 4, 5\} \cap \{1, 4\} = \{1, 4\}$
$[1]_a \cap [1]_b = \{1, 2, 4, 5\} \cap \{1, 2, 3\} = \{1, 2\}$

and only two of them

$[1]_b \cap [1]_c = \{1, 2, 3\} \cap \{1, 4\} = \{1\} \subseteq [1]_d = \{1, 2\}$
$[1]_a \cap [1]_b = \{1, 2, 4, 5\} \cap \{1, 2, 3\} = \{1, 2\} \subseteq [1]_d = \{1, 2\}$

are reducts of the family F. Hence, we have two value reducts: $b(1) = 0$ and $c(1) = 1$ or $a(1) = 1$ and $b(1) = 0$. This means that the attribute values of attributes a and b or c and b are characteristic for decision class 1 and do not occur in any other decision classes in the decision table. It can be noticed that the value of attribute b is the intersection of both value reducts, $b(1) = 0$, i.e., it is the core value.

In Table 3 value reducts for all decision rules are listed. It can be observed from the Table 3 that for decision rules 1 and 2 we have two values reducts of condition attributes. Decision rules 3, 4, and 5 have only one value reduct of condition attributes for each decision rule now. The remaining decision rules 6 and 7 contain two and three values reducts respectively. Considering one reduct for each of these decision rules we can have 24 (not necessarily distinct) solutions. One such solution is given in Table 4.

**Table 3**

| U | a | b | c | d |
|---|---|---|---|---|
| 1 | 1 | 0 | × | 1 |
| 1' | × | 0 | 1 | 1 |
| 2 | 1 | 0 | × | 1 |
| 2' | 1 | × | 0 | 1 |
| 3 | 0 | × | × | 0 |
| 4 | × | 1 | 1 | 0 |
| 5 | × | × | 2 | 2 |
| 6 | × | × | 2 | 2 |
| 6' | 2 | × | × | 2 |
| 7 | × | × | 2 | 2 |
| 7' | × | 2 | × | 2 |
| 7'' | 2 | × | × | 2 |

**Table 4**

| U | a | b | c | d |
|---|---|---|---|---|
| 1 | 1 | 0 | × | 1 |
| 2 | 1 | 0 | × | 1 |
| 3 | 0 | × | × | 0 |
| 4 | × | 1 | 1 | 0 |
| 5 | × | × | 2 | 2 |
| 6 | × | × | 2 | 2 |
| 7 | × | × | 2 | 2 |

From Table 4 it is evident that decision rules 1 and 2 are identical, and so are rules 5, 6, and 7, given in Table 5.

**Table 5**

| U | a | b | c | d |
|---|---|---|---|---|
| 1, 2 | 1 | 0 | × | 1 |
| 3 | 0 | × | × | 0 |
| 4 | × | 1 | 1 | 0 |
| 5, 6, 7 | × | × | 2 | 2 |

From Table 5, we get the final decision table, which is given as Table 6

**Table 6**

| U | a | b | c | d |
|---|---|---|---|---|
| 1 | 1 | 0 | × | 1 |
| 2 | 0 | × | × | 0 |
| 3 | × | 1 | 1 | 0 |
| 4 | × | × | 2 | 2 |

This is the reduced decision table termed as minimal. Here, out of seven, four decisions are selected and others are removed.

### 5.2 Algorithm for Feature Selection using Rough Sets

It evident from section 3.3 that after projection into the eigenspace the dimension of the input feature vectors is reduced Q (Q < n). Let us assume that after transformation of input image X, we get $X'$ i.e. $X' = W^T X$. Now, to reduce further we apply the following algorithm

**Step1.**  Obtain $X'_d$, which is a discrete form of $X'$.

**Step2.** Construct the decision table using $X'_d$ as condition and attribute value representing corresponding class as decision.
**Step3.** Find the respective core for each of the decision rules as discussed in 5.1.3 and redraw the table in the form of Table 2.
**Step4.** Find the value reduct as discussed in 5.1.3 and redraw the table in the form of Table 3.
**Step5.** Reduce the table as described by Table 4, Table 5, and Table 6.

The new representation of the faces is still very complex and variant. A method of pattern detection is required that is flexible and is able to recognize complex patterns. A neural network has been chosen specifically a multi-layer perceptron.

## 6. Multi-layer Perceptron

After feature reduction these feature vectors are classified using a multilayer perceptron. The network selected here is trained using back error propagation learning rule. The network consists of multiple layers of simple, two-state, sigmoid processing elements or neurons called nodes. It's principle of operation is very simple and due to this simplicity backpropagation have been used in more than 80% of the neural network applications. The back-propagation algorithm learns the weights for a fixed number of nodes and interconnections in the network. For every input vector the difference between the actual output, propagated from the input layer to the output layer, and desired output is back-propagated from output layer to the previous layers to update their interconnection weights. The initialization of weights determines the starting point in the error surface and effects the training process to end up in a local or global minimum. If the weights are initialized with larger values then the inputs to the activation function will be very high and due to that the activation function will operate in the saturation region and by which the changes in the weight values will be near to zero. Moreover, if weights are initialized to very small values then the learning process will become very slow and might not even converge. In this work, the weights are initialized with random values between -0.05 and 0.05 and updated according to the amount of error propagated backward.

Generally, the weight learning rule in a neural network can be given as the increment of the weight vector($w_{ij}$), from $j^{th}$ layer to $i^{th}$ layer, produced by the learning step at time t is proportional to the product of the learning signal r and the output vector generated by $i^{th}$ layer. This can be given as

$$\Delta w_{ij}^{(t)} \propto r \times x_j^{(t)} \text{ or } \Delta w_{ij}^{(t)} = \eta \times r \times x_j^{(t)} \quad (15)$$

where, $\eta$ is a positive constant called the learning coefficient that determines the rate of learning.

For the backpropagation learning rule, r is replaced by $\delta_i$, the amount of error back propagated from $i^{th}$ layer to $j^{th}$ layer neurons to update $w_{ij}$, which can be given as

$$w_{ij}^{(t+1)} = w_{ij}^{(t)} + \eta \delta_i x_j^{(t)} \quad (16)$$

The algorithm tries to minimize the squared error i.e. the sum of the squared differences of network output and target output for each of the feature vector used for training. The least mean square (LMS) error at the output level can be given as

$$E = \tfrac{1}{2} \sum_{i=1}^{n} (d_i - y_i)^2 \quad (17)$$

where, $d_i$ and $y_i$ are desired and actual output respectively for $i^{th}$ feature vector of a training set. The size of the training set is n. This error is thresholded by a tolerable limit($E_t$) rather than zero to complete training procedure early. If the network is trained with all the input vectors for one training set once then it is said as the completion of one epoch. A network converges if the error(E) proceeds towards $E_t$. Generally, a network converges with more number of epochs, if it is capable to separate different classes used during training. To design a neural network is not very easy. There is no formula available to find the architecture of a network like number of hidden layers, number of nodes in individual hidden layer and so on.

## 7. Experimental Results

For the investigation and comparison of results of different face recognition methods, a compound database[21] has been used. The compound database is composed of two face databases: one of them is the Cambridge (ORL) database containing 400 images of 40 distinct persons. Another database(D1) is created in our own laboratory, which includes 200 frontal views of 10 persons. Hence, in total, the compound database contains 600 images of 50 different individuals.

At first, experiments were conducted to study the face recognition performance i.e. whether recognition performance increases with the increase in number of eigenvectors in the eigenspace. The result is shown in the figure 2. A general trend is observed that after certain number of increase in the dimension there is no significant improvement in the performance. So, some of the eigenvector may be removed. After eigenspace projection, as discussed in section 3.3, the dimension of the feature vector has been reduced to 27. After further reduction (or through selection), as discussed in section 5.2, the dimension of the final feature vector has been computed as 9.

To classify them, as discussed in section 6, we have used a Multi-layer Perceptron. The network used in this work consists of three different layers. These are input, output, and one hidden layer. The size of the input layer is taken as nine to fit input feature vectors. Initially, experiments were conducted without any hidden layer but the network did not converge. Therefore, hidden layer had been introduced with an assumption that the given feature vectors are linearly non-separable. To decide about the number of nodes in hidden layer, several training experiments were conducted and finally it was decided as five.

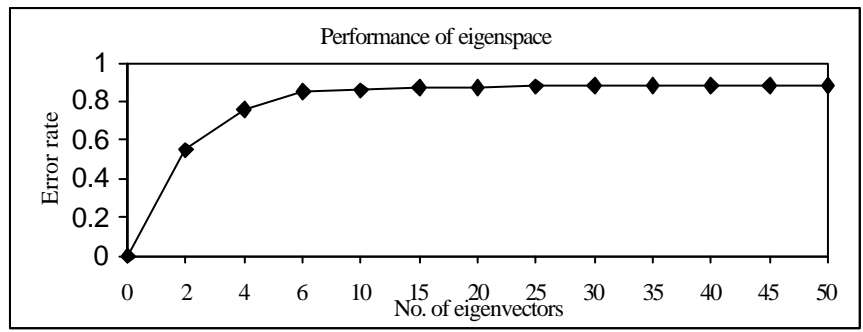

**Figure 2**

For these networks two-state output either true or false has been considered. As the output of any neurons in the network is a monotonic nonlinear (unipolar sigmoid) function, attainment of 0 (false) or 1 (for true) requires an infinite loop. Therefore, output values are actually distributed over a suitable interval [0.2, 0.8]. The experimental result shows that the performance of the classifier remains same with a significant improvement in the training time requirement in comparison to the result reported in [16].

## 8. Conclusion

In this paper we have presented a method to reduce the dimension of feature vector. The method, reported in this paper, has two steps. First step involves a technique for dimensionality reduction called eigenspace projection (also known as Principal Component Analysis). This method has already been used widely for dimension reduction and human face recognition. In the second step concept of 'reduct' and 'core' from rough set theory have been considered and this investigation has revealed the capability of rough set theory in feature selection and dimensionality reduction. This method has shown significant improvement in case of training time requirement i.e. the classifier network converges faster and the recognition rate has also increased.

**Acknowledgments**

Authors are thankful to the "Centre for Microprocessor Application for Training Education and Research" and "Project on Storage Retrieval and Understanding of Video for Multimedia, at the Department of Computer Science and Engineering, Jadavpur University, Kolkata - 700 032 and also Department of Computer Science and Engineering, University of Calcutta, Kolkata - 700 009 for providing the necessary facilities for carrying out this work.